\documentclass[twocolumn]{article}  
\usepackage{spconf,amsmath,graphicx,hyperref}
\usepackage{booktabs}
\usepackage{siunitx}
\usepackage{graphicx}
\usepackage{array}
\usepackage{multirow}
\usepackage{makecell}

\usepackage[x11names]{xcolor}  
\usepackage{caption}

\graphicspath{{Images/}}

\title{Are VLMs Ready for Lane Topology Awareness in Autonomous Driving?}
\name{
  \parbox{\textwidth}{\centering
  Xin Chen$^{1}$, Jia He$^{1}$, Maozheng Li$^{1}$, Dongliang Xu$^{1*}$, Tianyu Wang$^{2}$, \\
  Yixiao Chen$^{3}$, Zhixin Lin$^{1}$, Yue Yao$^{1*}$\thanks{*Corresponding authors. Code: \href{https://github.com/chenxin-sdu/TopoAware-Bench}{TopoAware-Bench}
}
  }
}
\address{$^1$Shandong University\quad
$^2$MBZUAI\quad $^3$Sems
}
\begin{document}
\ninept
\maketitle
\begin{abstract}


Vision-Language Models (VLMs) have recently shown remarkable progress in multimodal reasoning, yet their applications in autonomous driving remain limited. In particular, the ability to understand road topology, a key requirement for safe navigation, has received relatively little attention. While some recent works have begun to explore VLMs in driving contexts, their performance on topology reasoning is far from satisfactory. In this work, we systematically evaluate VLMs’ capabilities in road topology understanding. Specifically, multi-view images are projected into unified ground-plane coordinate system and fused into bird’s-eye-view (BEV) lanes. Based on these BEV lanes, we formulate four topology-related diagnostic VQA tasks, which together capture essential components of spatial topology reasoning. Through extensive evaluation, we find that while frontier closed-source models (\emph{e.g.,} GPT-4o) achieve relatively high accuracy in some tasks, they still fail in some spatial questions that humans can answer (\emph{e.g.,} GPT-4o achieves only 67.8\% in vector, a two-class classification problem). Furthermore, we find open-source VLMs, even at 30B scale, struggle significantly. These results indicate that spatial reasoning remains a fundamental bottleneck for current VLMs. We also find that the model’s capability is positively correlated with model size, length of reasoning tokens and shots provided as examples, showing direction for future research.
\end{abstract}
\begin{keywords}
Vision-Language Model, Lane Topology Awareness, Autonomous Driving
\end{keywords}
\section{Introduction}
\label{sec:intro}


A core requirement for autonomous driving is the ability to understand the lane topology, which refers to the structure and related properties of the driving environment. Unlike low-level sensing tasks, such as lane segmentation or object detection, lane topology awareness emphasizes reasoning about lane connectivity, intersection geometry, and relative directional relationships. These elements are indispensable for safe navigation and high-level decision making. Only when the model is accurate at the topological level can it lay a truly safe foundation for subsequent decision-making and trajectory planning.

VLMs are appealing for this setting. The latest generation of VLMs has shown great potential in multi-modal understanding, and its application scope has extended from general visual question answering to a variety of specific tasks. VLMS combine visual input with natural language reasoning, and this ability makes them strong candidates for autonomous driving~\cite{tian2024drivevlm,wang2025omnidrive,wang2023drivemlm,sima2024drivelm}. In autonomous driving, it is important to be able to reason about road conditions accurately. Furthermore, VLMs provide a unified framework that can reduce the fragmentation of task-specific modules, providing a scalable approach to integrate perception, reasoning, and interaction. These properties have driven attempts to apply VLMs to lane topology awareness for autonomous driving.

\begin{figure*}[!t]
\centering

\includegraphics[width=1\textwidth]{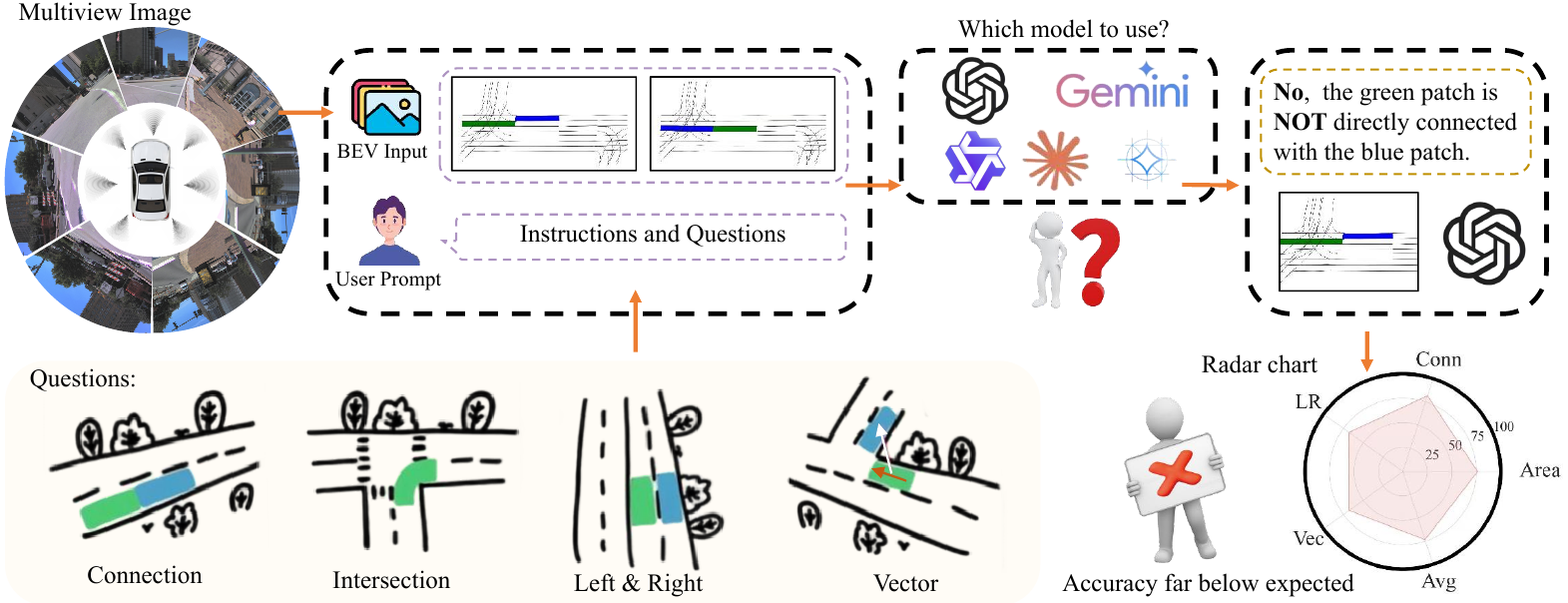}
\vspace{-2.1em}
\caption{Benchmark framework for evaluating the lane topology awareness capability of VLMs. BEV features are generated from multi-view images, different models are selected for inference, answers and results for four different subtasks are output, and the model performance on different tasks and different models are compared and evaluated through radar charts combined with user questions.}
\label{fig1:env}
\vspace{-1.7em}
\end{figure*}

However, we find that simply applying VLMs does not work well. Current VLMs are still limited when dealing with lane topology inference tasks. Most existing models are good at object-level recognition or descriptive description, but struggle with lane understanding. Without an explicit mechanism to represent and reason about graph like structures, VLMs often generate seemingly irrational responses. This gap highlights the urgent need for system evaluation and target benchmarks that can reveal the limitations of current VLMs for lane topology inference.

Specifically, we reuse and relabel the VQA tasks from Zhang \emph{et al.}~\cite{zhang2025chameleon} to evaluate the capabilities of VLMs for lane topology awareness. We introduce the Topology Awareness Benchmark for Lane Reasoning (TopoAware-Bench), which employs a lane segmentation model to process the reused multi-view images and extract road semantics. The extracted semantics are projected into a unified ground-plane coordinate system and fused into BEV lane representations, which are directly annotated as ground truth to isolate VLM reasoning performance from upstream perception errors. Based on these BEV lane graphs, we formulate four targeted VQA tasks—LeftRight, Intersection, Connection, and Vector Alignment—each probing a distinct aspect of spatial and relational reasoning. Together, these tasks capture the essential components of lane topology reasoning, providing a representative yet lightweight diagnostic tool for evaluating both closed-source and open-source VLMs. The evaluation benchmark framework is illustrated in Fig.~\ref{fig1:env}.

Experimentally, we find cutting-edge closed-source VLMS such as GPT-4o~\cite{hurst2024gpt} and Claude-3.5~\cite{anthropic2024model} achieve relatively strong results, but even large-scale open source models clearly underperform. These findings highlight that spatial reasoning is a fundamental bottleneck for the current generation of VLMS. By providing a standardized, interpretable, and domain-specific benchmark, TopoAware-Bench not only enables a reliable evaluation of lane topology awareness in autonomous driving, but also provides a foundation for advancing research on spatial reasoning more broadly, spanning graph learning, concrete AI, and geometry understanding.


\renewcommand{\arraystretch}{1}
\begin{table*}[t]
\footnotesize
\renewcommand{\arraystretch}{0.95}
\centering
\caption{Performance comparison of various vision language models on lane topology reasoning tasks. Metrics include Accuracy (Acc) and Recall for four subtasks: Intersection, Connection, LeftRight, and Vector, along with the overall Average Accuracy. Higher values indicate better performance.}
\vspace{-1em}
\setlength{\tabcolsep}{1mm}
\label{tab:model_performance}
\begin{tabular}{>{\centering\arraybackslash}m{1.1cm} m{3.0cm} c cc cc cc cc c}
\toprule
\textbf{Type} & \textbf{Model} & \textbf{\#Size} & \multicolumn{2}{c}{\textbf{Intersection}} & \multicolumn{2}{c}{\textbf{Connection}} & \multicolumn{2}{c}{\textbf{LeftRight}} & \multicolumn{2}{c}{\textbf{Vector}} & \textbf{Average} \\
& & & Acc (\%) & Recall (\%) & Acc (\%) & Recall (\%) & Acc (\%) & Recall (\%) & Acc (\%) & Recall (\%) & Acc (\%) \\

\midrule
-- & \textbf{random} & -- & 46.0 & 46.1 & 48.7 & 51.1 & 34.8 & 35.3 & 54.4 & 51.5 & 46.0 \\
\midrule

\multirow{3}{*}[1ex]{%
\rotatebox{90}{%
    \parbox[c]{4.5em}{\centering Closed\\[-1pt]Source}%
  }%
}& GPT-4o \cite{hurst2024gpt} & -- & \textbf{\boldmath 76.4} & 91.6 & 81.5 & 75.2 & 68.0 & 69.8 & 67.8 & 50.0 & \textbf{\boldmath 73.5} \\[2pt]
& Claude-3.5-Sonnet \cite{anthropic2024model} & -- & 57.6 & 19.5 & 76.2 & 51.1 & 67.4 & 65.9 & 62.9 & 25.8 & 66.0 \\
& Gemini-2.5-Flash \cite{comanici2025gemini} & -- & 68.3 & 64.9 & \textbf{83.9} & 94.9 & \textbf{\boldmath 75.1} & \textbf{\boldmath 76.2} & 53.4 & 7.6 & 70.2 \\
\midrule

\multirow{14}{*}{}
& \multirow{3}{*}{InternVL3 \cite{zhu2025internvl3}} & 2B & 49.5 & \textbf{\boldmath 94.8} & 52.0 & 59.9 & 40.4 & 41.5 & 46.6 & 93.2 & 47.2 \\
& & 8B & 54.1 & 88.9 & 51.3 & 2.9 & 42.2 & 39.1 & 51.9 & 68.9 & 49.9 \\
& & 14B & 64.4 & 48.1 & 53.4 & 6.6 & 55.9 & 57.0 & 49.8 & 22.7 & 55.9 \\
\cmidrule(l){2-12}
\multirow{14}{*}{\rotatebox{90}{\parbox[c]{2.8em}{\centering Open\\[-1pt]Source}}}& \multirow{2}{*}{LLaVA-v1.5 \cite{liu2024improved}} & 7B & 48.2 & 77.9 & 48.7 & \textbf{\boldmath 95.6} & 48.5 & 48.9 & 45.6 & 96.2 & 47.7 \\
& & 13B & 56.3 & 79.9 & 51.3 & 48.9 & 52.4 & 53.5 & 48.1 & 59.1 & 52.0 \\
\cmidrule(l){2-12}
& Qwen2-VL \cite{wang2024qwen2} & 2B & 50.8 & 93.5 & 52.0 & 44.5 & 40.2 & 41.3 & 45.2 & 47.7 & 47.1 \\
\cmidrule(l){2-12}
& \multirow{3}{*}{Qwen2.5-VL \cite{bai2025qwen2}} & 3B & 59.6 & 53.3 & 53.7 & 29.2 & 57.2 & 58.8 & 49.5 & 81.8 & 55.0 \\
& & 7B & 54.1 & 29.2 & 57.0 & 33.6 & 60.9 & 62.9 & \textbf{\boldmath 71.7} & 86.4 & 60.9 \\
& & 32B & 63.8 & 66.9 & 71.5 & 78.8 & 66.0 & 68.5 & 65.7 & 37.9 & 66.8 \\
\cmidrule(l){2-12}
& Kimi-VL \cite{team2025kimi} & 3B & 54.1 & 88.3 & 54.7 & 9.5 & 42.0 & 39.1 & 55.5 & 75.8 & 51.5 \\
\cmidrule(l){2-12}
& SmolVLM \cite{marafioti2025smolvlm} & 2.2B & 46.9 & 50.0 & 47.7 & 53.3 & 36.7 & 34.2 & 53.4 & 83.3 & 46.2 \\
\cmidrule(l){2-12}
& Phi-3.5-vision \cite{abdin2024phi} & 3.5B & 50.2 & 12.3 & 54.0 & 3.6 & 38.5 & 34.9 & 51.9 & 59.9 & 48.7 \\
\cmidrule(l){2-12}
& Gemma-3 \cite{team2025gemma} & 27B & 64.7 & 29.9 & 63.8 & 80.3 & 56.5 & 56.8 & 56.2 & 12.9 & 60.0 \\
\cmidrule(l){2-12}
& MiMo-VL \cite{coreteam2025mimovltechnicalreport} & 7B & 63.1 & 81.8 & 56.0 & 91.9 & 62.3 & 66.3 & 57.2 & 83.3 & 59.7 \\
\cmidrule(l){2-12}
& Pixtral \cite{agrawal2024pixtral} & 12B & 49.8 & 79.2 & 48.0 & 94.2 & 47.4 & 46.0 & 47.4 & \textbf{\boldmath 97.7} & 48.2 \\
\cmidrule(l){2-12}
& Mistral-Small-3.1 \cite{mistral2025small3} & 24B & 53.1 & 6.5 & 57.1 & 18.3 & 57.8 & 59.7 & 51.2 & 81.1 & 54.8 \\
\midrule
\addlinespace[5pt]
\multirow{2}{*}[1ex]{%
\rotatebox{90}{%
    \parbox[c]{3.2em}{\centering Driving\\Specific}%
  }%
}
& RoboTron-Drive \cite{huang2025robotron} & 7B & 51.5 & 33.8 & 60.0 & 46.0 & 42.8 & 40.6 & 47.0 & 84.1 & 50.3 \\[2pt]
& Dolphins \cite{ma2024dolphins} & 7B & 51.5 & 28.6 & 54.4 & 14.6 & 34.3 & 32.6 & 48.0 & 12.9 & 47.1 \\
\addlinespace[2pt]
\bottomrule

\end{tabular}
\end{table*} 
\begin{table}[h]
\centering
\caption{Ablation study of inference strategies and prompt components. On the basis of InternVL3-8B model, the accuracy of four sub-tasks under different reasoning strategies are analyzed.}  
\vspace{-1em}
\resizebox{\columnwidth}{!}{%
\begin{tabular}{lccccc}
\toprule
Method & Area (\%) & Conn (\%) & LR (\%) & Vec (\%) & Avg (\%) \\
\midrule
InternVL3-8B & 54.1 & 51.3 & 42.2 & 51.9 & 49.9 \\
+TTS & 54.7 & 52.0 & 43.9 & 53.0 & 50.9 \\
+TTS+few shots & 56.6 & 53.3 & 44.1 & 53.7 & 51.9 \\
\bottomrule
\end{tabular}%
}
\label{tab:TTS}
\vspace{-2em}
\end{table}

\section{Related Work}

\textbf{Lane Topology Awareness in Autonomous Driving.}
In autonomous driving, lane topology perception is a basic prerequisite for safety decision-making and path planning. 
Recent studies have improved lane topology perception in several ways. Li \emph{et al.}~\cite{yang2024sstp} reuse the attention mechanism to achieve synchronous inference between detection and topology, and introduce knowledge distillation to improve the performance in mapless scenarios. Fu \emph{et al.}~\cite{fu2025topopoint} significantly improve the accuracy of topology inference by explicitly modeling lane endpoints. In addition, SDMap is introduced to improve the inference performance~\cite{luo2024augmenting}, and TopoLogic~\cite{fu2024topologic} combines the distance and semantic similarity to reduce false connections. Although the existing methods have made progress in feature reuse, endpoint modeling and map priors, they still have common disadvantages. On the one hand, high-quality topology annotation data is scarce. On the other hand, most methods lack explicit modeling of the 3D topology and struggle to maintain robustness in terms of geometric consistency and orientation inference.

\textbf{VLMs and VLMs' Spatial Understanding.}
VLMs have shown unique advantages in multimodal understanding~\cite{hurst2024gpt,anthropic2024model,comanici2025gemini,zhu2025internvl3,liu2024improved,wang2024qwen2,bai2025qwen2,team2025kimi,abdin2024phi,coreteam2025mimovltechnicalreport,agrawal2024pixtral,mistral2025small3}. Representative models such as GPT-4o ~\cite{hurst2024gpt} and Gemini ~\cite{comanici2025gemini} perform well in visual question answering and cross-modal reasoning, but their application in autonomous driving scenarios is still limited ~\cite{xu2025trajectory,xie2025vlms}, especially in the key link of lane topology perception. The performance of VLMs in this aspect still lacks a scientific systematic evaluation framework. Existing benchmarks focus more on detection accuracy or single-task performance, and cannot fully quantify the performance of VLMs on key tasks such as spatial structure modeling, connectivity inference, and intersection judgment. To this end, this paper proposes TopoAware-Bench, a VLM spatial reasoning benchmark program for autonomous driving scenarios, which aims to systematically evaluate the topological reasoning ability of VLMs on real road data and provide a reference for subsequent research.

\label{sec:format}

\section{Benchmark Construction}
We reuse and manually relabel data from Zhang \emph{et al.}~\cite{zhang2025chameleon} and propose a new benchmark, TopoAware-Bench, a standardized benchmark for lane topology reasoning in autonomous driving. TopoAware-Bench is designed as a structured, interpretable, and unbiased diagnostic tool, with multi-view inputs including a BEV and a front perspective view (PV), which explicitly test models’ ability to integrate different geometric representations. To construct the benchmark, we reuse and relabel the 1,300 VQA questions from Zhang \emph{et al.}~\cite{zhang2025chameleon}, and package them into a consistent and reproducible framework. On average, each task contains 300 curated samples, each consisting of a visual prompt and a textual query. TopoAware-Bench is suitable for evaluating both open-source and closed-source models of varying scales (from 2B to 30B+ parameters), providing a standardized and reliable foundation for advancing research on spatial reasoning in autonomous driving.

Visual prompts are generated from both the BEV and the PV, which respectively capture geometric structures and semantic context. Textual queries are formulated as concise questions emphasizing semantic and spatial reasoning. Each sample in TopoAware-Bench therefore consists of visual inputs $x_v = \{ I_{\text{BEV}}, I_{\text{PV}} \}$, where $I_{\text{BEV}}$ encodes lane-level topology and geometry and $I_{\text{PV}}$ preserves local semantic context; textual queries $x_t$, instantiated from standardized templates to serve as reasoning triggers; and task-specific labels $y$, annotated for the four subtasks of TopoAware-Bench (Intersection, Connection, LeftRight, and Vector). This dual-view design enables us to examine models’ ability to integrate multi-view geometric information in a consistent and structured way.
To ensure fair evaluation, we rigorously balanced the dataset. For binary classification tasks (Intersection, Connection, Vector), the ratio of positive to negative samples is approximately $1\!:\!1$. For LeftRight, left and right cases are symmetrically distributed. We also provide detailed metadata (\emph{e.g.}, color highlights, lane IDs, arrow orientations) to ensure reproducibility. 

\textbf{Task Definition.}
TopoAware-Bench is organized around four subtasks, each probing a distinct aspect of spatial and semantic reasoning:
\[
\mathcal{T}_{\text{TopoAware}} = \{\mathcal{T}_{\rm intersect}, \mathcal{T}_{\rm conn}, \mathcal{T}_{\rm LR}, \mathcal{T}_{\rm vec}\},
\]
where each subtask corresponds to a reasoning function
\begin{equation}
    y_i = f_{\theta,i}(I_{\text{BEV}}, I_{\text{PV}}, q_i), \quad i \in \{1,2,3,4\},
\end{equation}
with $q_i$ denoting the standardized textual query and $y_i$ the ground-truth label. Specifically, the Intersection subtask examines whether a highlighted lane segment lies within the spatial extent of an intersection region, reflecting the model’s ability to reason about global contextual information. The Connection subtask evaluates local structural continuity by determining whether two lane segments belong to the same lane and are end-to-end adjacent. The LeftRight subtask addresses relative spatial positioning by requiring the model to judge whether one lane segment is located to the left or right of another in the BEV representation. The Vector subtask tests directional consistency by comparing the orientation of two directional arrows and assessing whether their headings are aligned within a specified angular threshold. 

\textbf{Prompt Design.}
Each TopoAware-Bench subtask is paired with a standardized prompt template to ensure evaluation is precise and reproducible. Visual prompts highlight relevant lane segments in BEV and PV views, while textual queries are phrased in consistent and unambiguous forms.
For example, the standardized prompt for the Connection subtask is: ``The black lines in the photos are lane boundaries. Two segments in different lanes don't have any connection relationship. Only two segments in the same lane end to end adjacent are considered as directly connected. You are an expert in determining adjacent lane segments in the image. Let's determine if the green segment is directly connected with the blue segmemt. Please reply in a brief sentence starting with `Yes' or `No'.''





\section{Experiment}
\begin{figure*}
\centering
\includegraphics[width=1\textwidth]{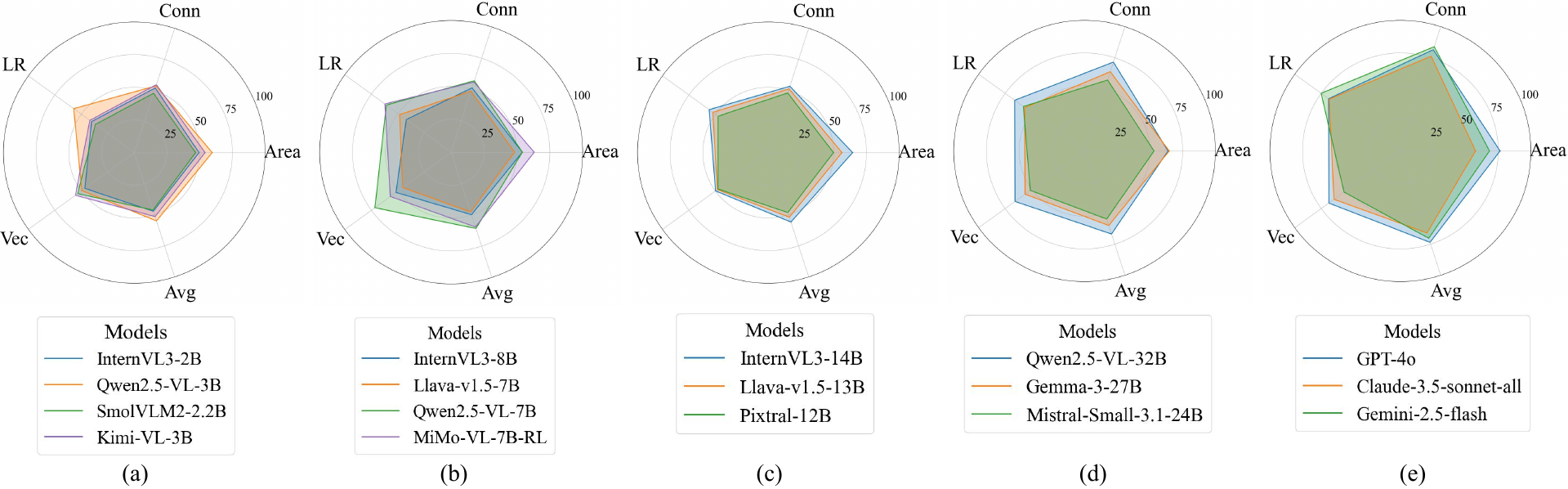}
\vspace{-2em}
\caption{Radar charts of model performance on topology awareness tasks grouped by parameter scale.  These radar charts illustrate the performance of different Vision-Language Models on the TopoAware-Bench, organized by parameter scale.  Tasks include Connection (Conn), LeftRight (LR), Intersection (Area), and Vector (Vec), with average performance (Avg) also shown. Subgraphs (a) to (e) compare models in different parameter ranges, including open source and closed source VLMs.}
\vspace{-1em}
\label{fig2:env}
\end{figure*}

\begin{figure}
\centering
\includegraphics[width=0.5\textwidth]{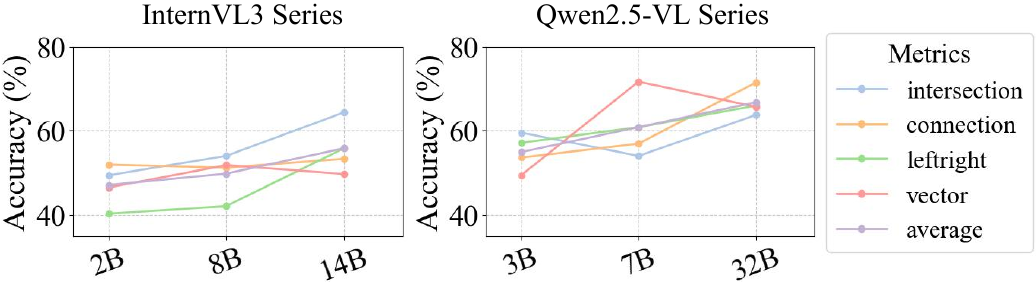}
\vspace{-1.9em}
\caption{Performance comparison of the same family of models with different parameter scales on the TopoAware-Bench task. The three parameter specifications of InternVL3 and Qwen2.5-VL series models are compared, and five evaluation dimensions are corresponding to different colors.}
\vspace{-1.5em}
\label{fig3:env}
\end{figure}

\textbf{Frontier closed-source models achieve relatively high accuracy but still fail on some specific spatial and geometric reasoning.}
The experimental results in Table~\ref{tab:model_performance} show significant differences in the performance of different models on the four categories of VQA subtasks. We find that while frontier closed-source models such as GPT-4o~\cite{hurst2024gpt}, Claude-3.5~\cite{anthropic2024model}, and Gemini-2.5~\cite{comanici2025gemini} achieve relatively high accuracy in some tasks, they still fail in certain spatial or geometric reasoning questions that humans can easily answer. For instance, GPT-4o~\cite{hurst2024gpt} reaches an average accuracy of 73.5\%, especially excelling in the Intersection (76.4\%) and Connection (81.5\%) subtasks, yet it still shows bottlenecks in vector alignment (67.8\%) tasks which is a two class classification problem. This indicates that even top-tier proprietary models remain imperfect in spatial reasoning, and that domain specialization alone does not provide clear advantages for topology-oriented tasks.


\textbf{In contrast, open-source VLMs show substantial limitations.}
Even at scales up to 30B parameters, models such as InternVL3 and LLaVA-1.5 struggle to exceed 52\% accuracy on average, with particularly weak performance in Connection and Vector reasoning tasks. In addition, recall-oriented evaluation further highlights the unreliability of open-source VLMs. For instance, InternVL3-8B-Instruct achieves 51.3\% accuracy on the Connection task but only 2.9\% recall, indicating near-total failure in retrieving true positives. Similarly, Phi-3.5-vision-instruct reaches just 12.3\% recall on Intersection and 3.6\% recall on Connection, performing worse than the random baseline. Even at larger scales, Gemma-3-27B demonstrates only 12.9\% recall in the Vector task, reflecting severe deficiencies in directional reasoning. These results suggest that although open-source models possess certain visual-language alignment capabilities, they still exhibit substantial deficiencies in demanding scenarios such as lane connectivity inference and directional alignment, which are critical for autonomous driving safety.

\begin{figure}
\centering
\includegraphics[width=0.4\textwidth]{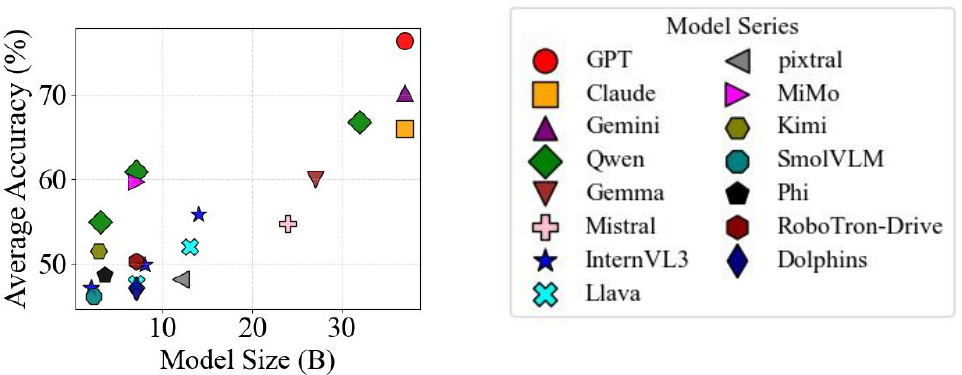}
\vspace{-1em}
\caption{Scatter plot showing the relationship between model size (in billions of parameters) and average accuracy (\%) on the TopoAware-Bench. Different model series are represented with distinct markers.}
\label{fig4:env}
\vspace{-1em}
\end{figure}

\textbf{A clear trend emerging from the experiments is the strong positive correlation between model size and reasoning capability.}
Fig. \ref{fig3:env} illustrates the performance variation within two open-source model families (InternVL3 and Qwen2.5-VL). Both series show a clear scaling law: larger parameter scales consistently lead to higher accuracy. Fig. \ref{fig2:env} further visualizes this trend via radar plots, where low parameter models have smaller ``coverage'' and lower metric values, while high parameter models display wider ``coverage'' and stronger performance across subtasks. Finally, Fig. \ref{fig4:env} confirms this positive relationship at a macro level, showing that as parameter count increases from under 10B to over 30B, the average accuracy steadily rises from the 40–50\% range to the 60–70\% range.

\begin{figure}
\centering
\includegraphics[width=0.5\textwidth]{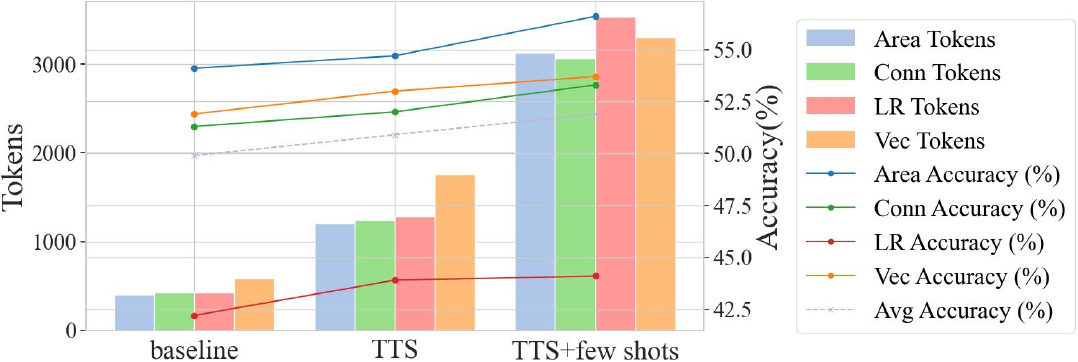}
\vspace{-2em}
\caption{The influence of test time scaling measured by thinking tokens and few shot samples. The left vertical axis represents the number of tokens, the right vertical axis represents the accuracy, and the horizontal axis represents the three methods baseline, TTS, and TTS + few shots in turn. Different color bars correspond to the number of Tokens in four tasks. The marked broken lines represent the accuracy of each of the four categories of tasks.}
\label{fig5:env}
\vspace{-1.5em}
\end{figure}

\textbf{Test time scaling~\cite{muennighoff2025s1} (length of reasoning tokens) has positive influence to model performance.}
As can be seen in Fig.~\ref{fig5:env}, our analysis shows that models producing longer chains of reasoning tend to achieve higher accuracy, especially in the Intersection and Vector subtasks. This suggests that the ability to sustain coherent multi-step reasoning, rather than just model size, contributes to performance gains. At the same time, excessively verbose reasoning without structural precision does not guarantee correctness, revealing the need for more efficient reasoning strategies.

\textbf{Finally, we investigate the role of specific task inference strategies through an ablation study on InternVL3-8B.}
In order to clarify the supplementary influence of reasoning strategies, we conducted an ablation study based on InternVL3-8B to examine the roles of Test Time Scaling (TTS) ~\cite{muennighoff2025s1} and few-shot examples. As shown in Table \ref{tab:TTS}, introducing TTS alone improved the average accuracy on the four subtasks by about 1\%. When TTS was combined with a few shots, performance increased further to 52\%. Together, these findings show that while parameter scale dominates raw performance, carefully designed reasoning strategies, both in terms of prompting and demonstration, offer an orthogonal path toward improving VLMs in lane topology inference.

\section{Conclusion}
In this work, we introduce TopoAware-Bench, a diagnostic benchmark with four structured VQA tasks to evaluate vision-language models on lane topology awareness. Experiments show a clear gap between closed-source models, which perform relatively well in intersection detection and connectivity, and open-source models, which struggle across tasks, especially in vector alignment. These findings reveal spatial reasoning as a core bottleneck for current VLMs and highlight the need for stronger geometric biases and training strategies, with TopoAware-Bench offering a standardized framework to drive progress in lane topology awareness.

\vfill\pagebreak
\section{Acknowledgment}
This work was supported by the Key Research and Development Program of Shandong Province, China (2025CXGC010901), the Shandong Province Overseas Young Talents Program, and the Shandong Key Laboratory of Industrial Network Security (2025CXPT069).


\begingroup
\ninept 
\setlength{\itemsep}{0pt}   
\setlength{\parskip}{0pt}   
\bibliographystyle{IEEEtran}
\bibliography{strings,refs}
\endgroup

\end{document}